# Exploiting the Power of Levenberg-Marquardt Optimizer with Anomaly Detection in Time Series

*Wenyi Wang[1], John Taylor[1,2,3] and Biswajit Bala[1]*

[1]Defence Science and Technology Group, Australia
[2]CSIRO, Data61, Australia
[3]Australian National University, School of Computing

## Abstract

The Levenberg-Marquardt (LM) optimization algorithm has been widely used for solving machine learning problems. Literature reviews have shown that the LM can be very powerful and effective on moderate function approximation problems when the number of weights in the network is not more than a couple of hundred. In contrast, the LM does not seem to perform as well when dealing with pattern recognition or classification problems, and inefficient when networks become large (e.g. with more than 500 weights). In this paper, we exploit the true power of LM algorithm using some real-world aircraft datasets. On these datasets most other commonly used optimizers are unable to detect the anomalies caused by the changing conditions of the aircraft engine. The challenging nature of the datasets are the abrupt changes in the time series data. We find that the LM optimizer has a much better ability to approximate abrupt changes and detect anomalies than other optimizers. We compare the performance, in addressing this anomaly/change detection problem, of the LM and several other optimizers. We assess the relative performance based on a range of measures including network complexity (i.e. number of weights), fitting accuracy, over fitting, training time, use of GPU's and memory requirement etc. We also discuss the issue of robust LM implementation in MATLAB and TensorFlow for promoting more popular usage of the LM algorithm and potential use of LM optimizer for large-scale problems.

## 1. Introduction

In applying machine learning (ML) to detecting faults of fielded machinery, an overwhelming challenge is the lack of fault data for supervised learning [1, 2]. Therefore, it is crucial to be able to detect any fault-induced anomaly based on unsupervised learning, where the function approximation can be a key task. The Levenberg-Marquardt (LM) algorithm is a well-known nonlinear optimization algorithm for function approximation. There are numerous papers in open literature about the LM algorithm and its applications. Roweis [3] gave an excellent intuitive description of the algorithm. It summarized that the algorithm works extremely well in practice and has become a virtual standard for optimization of medium sized nonlinear models (a few hundred weights for instance) – much faster than stochastic gradient descent plus momentum (SGDM) or adaptive momentum (ADAM). Mathworks published a technical note [4] that compared LM with other commonly used optimization algorithms using a number of test datasets. It concluded that the Levenberg-Marquardt algorithm generally has the fastest convergence with lower mean square errors than any of the other algorithms tested in many cases on function approximation problems for networks that contain up to a few hundred weights. This is especially advantageous when very accurate training is required. However, the advantage of LM can decrease as the number of weights in the network increases, and the LM performance is recognized as being relatively poor on pattern recognition problems. The memory storage requirements of the LM algorithm are larger than the other tested algorithms.

Neural networks (NN) can be very effective in fitting complex functions, but is very dependent on the choice of optimizers. As an example, we compare the LM algorithm with the popular ADAM optimizers by fitting a '*sinc*' function using MATLAB's neural network curve fitting tool – *nftool*. We use a single layer Dense (feedforward) model with 20 neurons and 100 epochs. In Figure 1 the NN model trained using the LM optimizer performs much better than by the ADAM optimizer.

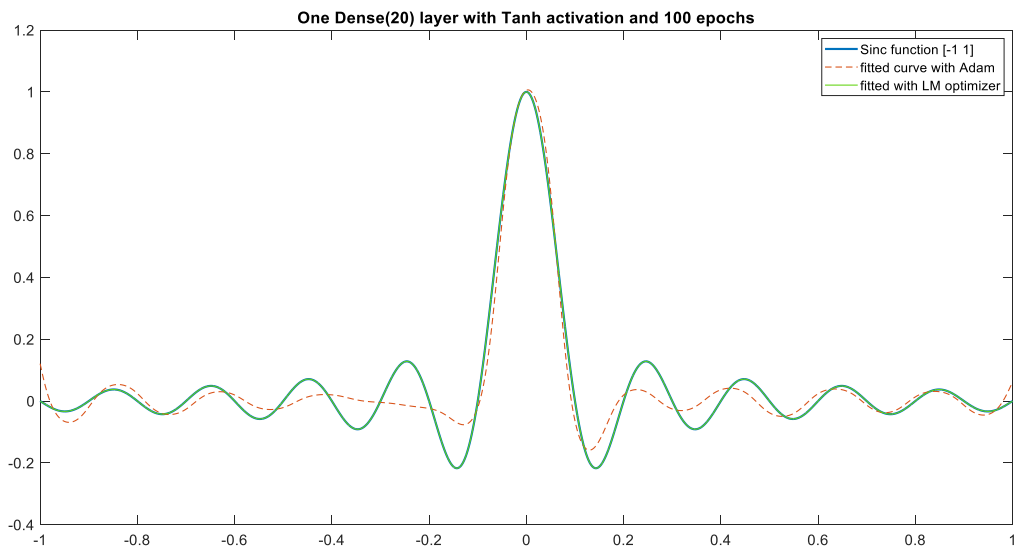

**Figure 1.** A simple comparison of optimizers using MATLAB NN curve fitting tool (note the blue and green curves are overlapped)

On dealing with a challenging practical problem of anomaly detection, we have trialed many different NN models including using LSTM, autoencoder, 1D convolutional NN and GANs. We found that none of these models with less than 200 hidden units seems to perform well under the ADAM optimizer. However, once we started to use the LM optimizer, a new horizon appeared with the simplest NN architecture – a single dense layer NN. We present some of the most interesting findings as a comparative study of LM with other optimizers.

We will use an aircraft engine failure dataset which contains 32 sequential events of engine operations. Figure 2*a* and 2*b* respectively show the engine's normalized speed and normalized vibration level with the end of each event being marked by the green triangles on the *x*-axis. When the engine failed in the last event, there was a very large spike (3.56) in the vibration data (Figure 2*b*). For reasons that are beyond the scope of this paper, there is high confidence that the data contain anomalies prior to the failure. We are tasked to detect any possible anomalies prior to the event of engine failure. As can be seen, these sequential event signals contain many abrupt changes or sudden jumps/dips, which forms a very challenging problem of function approximation or anomaly detection. What we aim to achieve is to accurately approximate the aircraft vibration signal in the training data using a NN model with engine speed as the input signal. The approximation needs to be robust enough and generalizable outside the training data, i.e. to avoid overfitting. We choose the first 15 events as the training/validation data, and the rest as the test data. The training and test data are separated by a red vertical dashed line as shown in Figure 2.

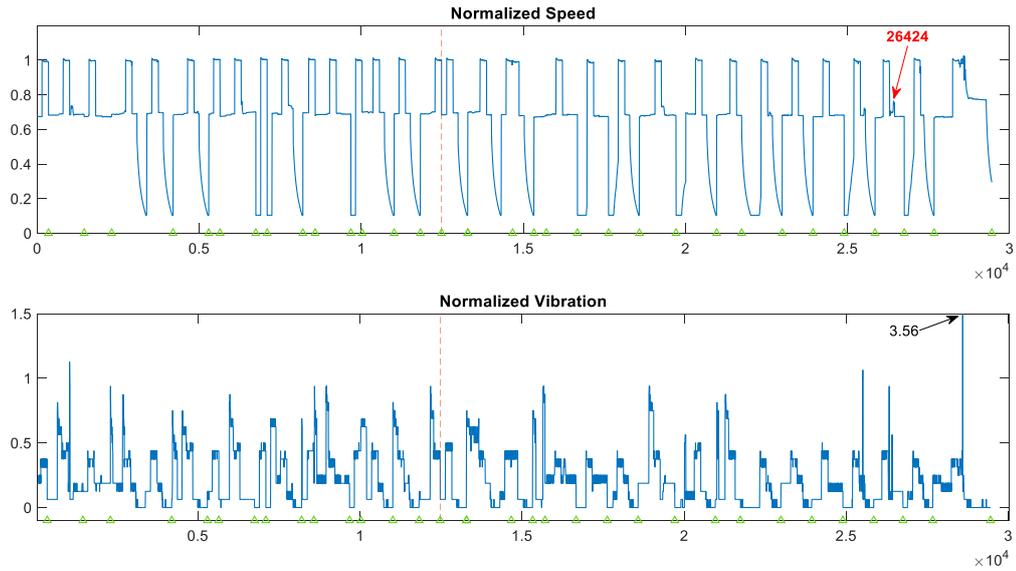

**Figure 2.** Normalized input and output signals

## 2. Anomaly detection in MATLAB

For shallow NN models, MATLAB has implemented the LM optimizer with the training function of '*trainlm*' as the default optimizer, but it is only available for CPUs. The LM optimizer is also not available for deep learning models in MATLAB.

### *2.1 Result with LM optimizer*

We use a single layer of dense model with 40 hidden neurons and the '*tanh*' transfer or activation function, the train-validation split is 60-40 with random division for each epoch. The model is trained in 300 epochs where no early stopping occurred with a validation tolerance of 3 epochs. The best performances (in mean squared error – '*mse*') on training and validation data were 0.0075 and 0.0078 respectively, and training time is 25 sec on a Predator Helios 300 CPU (note that GPU training is not currently available with the MATLAB version of the LM optimizer).

The beauty of this single-dense-layer-only structure is that we almost only need to choose the number of hidden neurons and maximum epochs. We found the combination of 40-neuron architecture plus a 300-epoch training regime can produce a well-balanced result, i.e. good performance (low MSE value) with minimal overfitting. At the same time, more neurons and epochs does not guarantee a significantly better result while carrying a much high risk of overfitting. Results in Figure 3 show that the model trained for 300 epochs is appropriately fitted with minimal overfitting as the prediction errors in the training data are comparable to those in the early part of the test data. We can also clearly see two strong anomalies (spikes) being detected at the two events (use the green triangles on the *x*-axis as location references) prior to the engine failure. For the purpose of this paper, we see anomalies as those prediction errors for the test data that have much bigger amplitudes than the biggest prediction error for the training data. However, this is a measure of relativity within each figure shown here, which is why the figures are not displayed with the same vertical range.

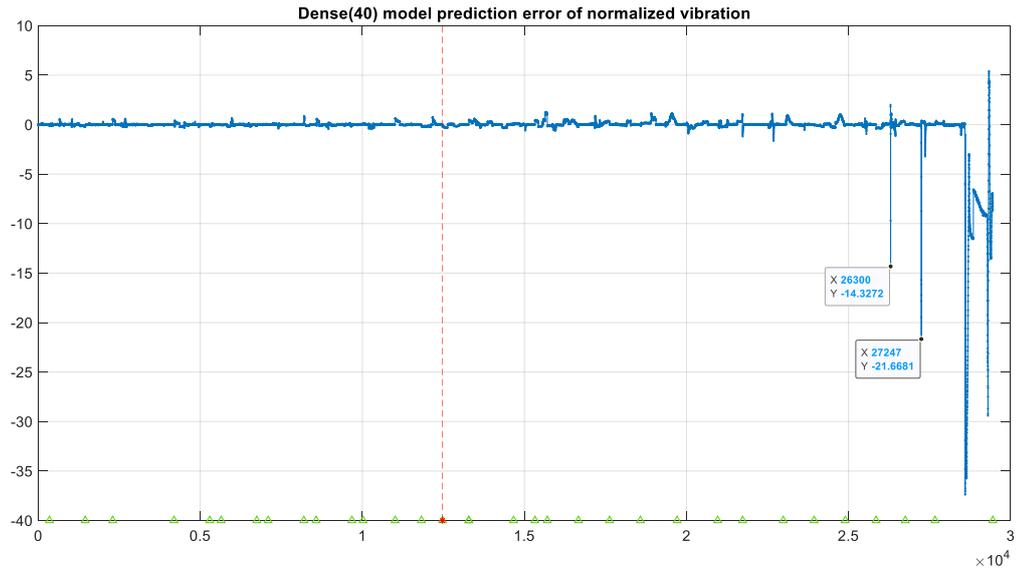

**Figure 3.** Prediction error of output vibration signal using a Dense (40 hidden units) NN model with the LM optimizer (maximum 300 epochs)

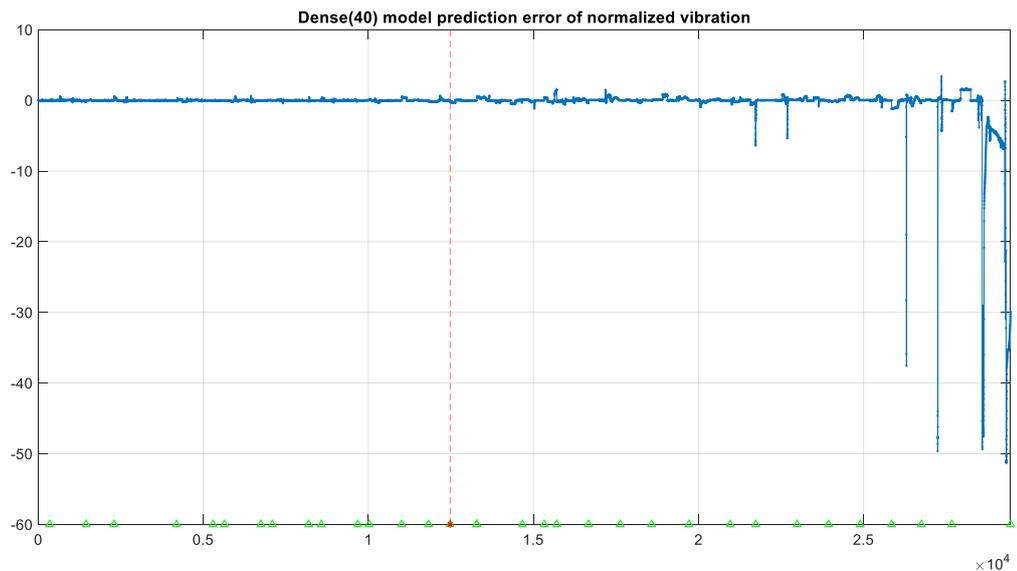

**Figure 4.** Prediction error of output vibration signal using a Dense (40 hidden units) NN model with the LM optimizer (maximum 1000 epochs)

Figure 4 shows the LM result with 1000 epochs, where the model is probably overfitted because the prediction errors in the early part of the test data are bigger than those in the training data. The two spikes between 20000 and 25000 sampling points are likely to be numerical artefacts. Numerical artefacts can come and go, and show up in different places (most likely with a much smaller amplitude) if we run the training multiple times. In other words, artefacts are of random nature. If we can see the same anomalies in the same places, i.e. last two events prior to failure, by multiple trained models and not in other places, we should be more certain that they are anomalies rather than artefacts.

*2.2 Results with other optimizers*

There are many optimizers available to shallow NN models in MATLAB. After many trials of different optimizers and network structures, we found that most optimizers are incapable of detecting the anomalies. These optimizers are unable to approximate the sudden and sharp changes in the signals. The best result with these optimizers is shown in Figure 5. It is produced by a three-layer dense model with [100,50,25] hidden neurons in the respective layers, and the resilient backpropagation (RP) optimizer and the '*tanh*' transfer function. The train-validation split is also 60-40 with random division for each epoch, the model is trained in 1000 epochs where no early stop occurred at the validation tolerance of 10 epochs. The performances (in mean squared error – '*mse*') on both training and validation data are 0.0074, and training time is 16 sec on a Predator Helios 300 GPU (GTX1060) training. Comparing this best result by other optimizers to the result by LM optimizer shown in Figure 3, we can see that the LM result is much more capable of detecting the anomalies prior to the engine failure with much higher relative magnitudes for the detected anomalies.

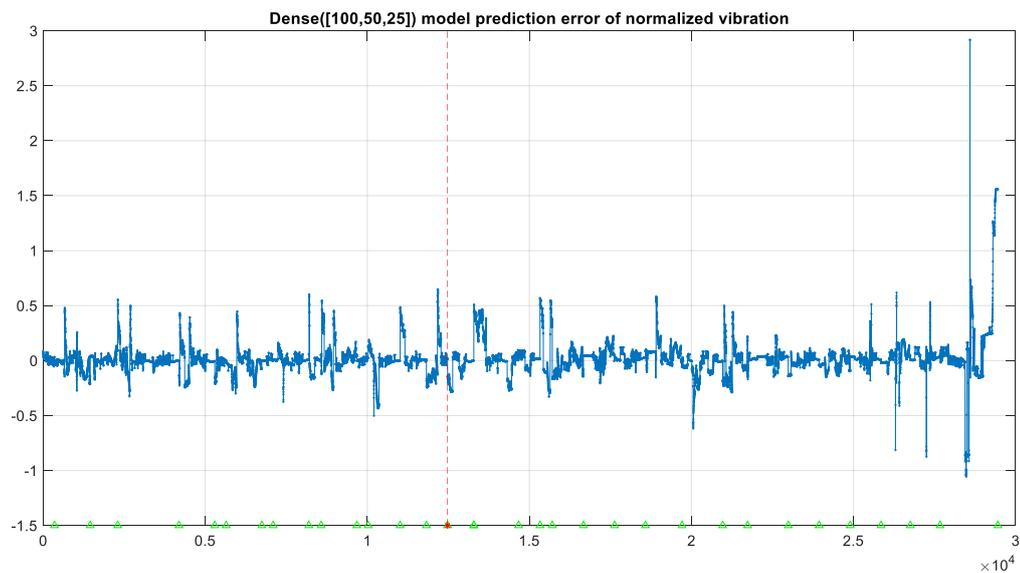

**Figure 5.** Prediction error of output vibration signal using a Dense([100,50,25]) NN model with the RP optimizer (maximum 1000 epochs)

## 3. Anomaly detection in TensorFlow-Keras

Because there is no built-in implementation of LM in Tensorflow and Keras, we imported a user contributed code for Levenberg-Marquardt (LM) algorithm by Fabio Di Marco from GitHub [5]. The LM code has been implemented so that the LM optimizer can be used for a GPU calculation with deep learning models. Figure 6 below shows the result of an autoencoder model with a single dense layer model of only 10 hidden neurons for both the encoder and decoder part of the model. The model is trained in 50 epochs using the LM optimizer and mean absolute error as loss function. From the model prediction error, we can clearly see the detected anomalies prior to the engine failure, but the model seems to have a relatively big prediction error (a spike) just before the 5000-sample point, which is likely to be a numerical artefact. The use of more hidden neurons and epochs doesn't seem to improve the performance substantially. In comparison, under the same model architecture but with ADAM optimizer, see Figure 7, we need many more hidden units (512 vs 10) and epochs (500 vs 50) to achieve a similar performance to the LM optimizer in Figure 6.

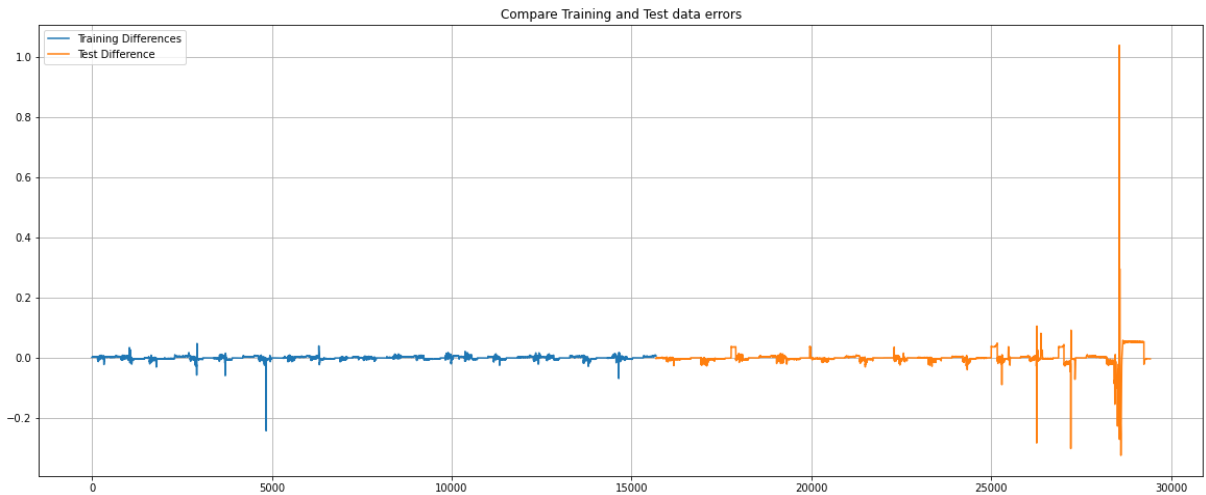

**Figure 6.** Prediction error of output vibration signal using an autoencoder-dense(10) NN model with LM optimizer (maximum 50 epochs)

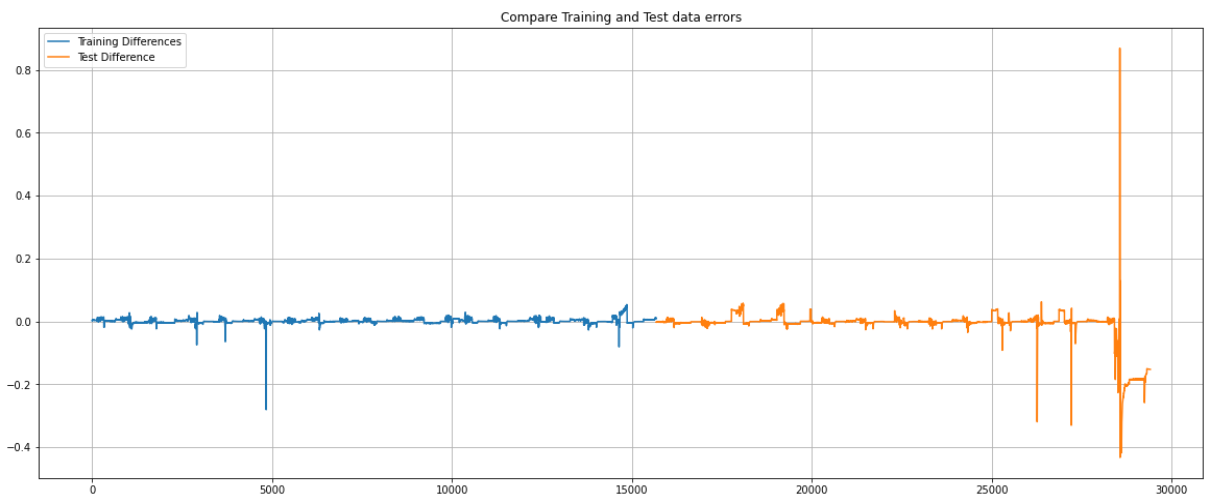

**Figure 7.** Prediction error of output vibration signal using an autoencoder-dense(512) NN model with ADAM optimizer (maximum 500 epochs)

With a 1D convolutional model (Conv1D) and ADAM optimizer, it seems that no matter how many hidden units and epochs we choose, we cannot achieve anything close to what Figure 6 shows. The best result is given in Figure 8, where the model has 5 Conv1D layers and over 2 million trainable weights and is trained over 500 epochs. The result shows evidence of overfitting with poor fit to test data.

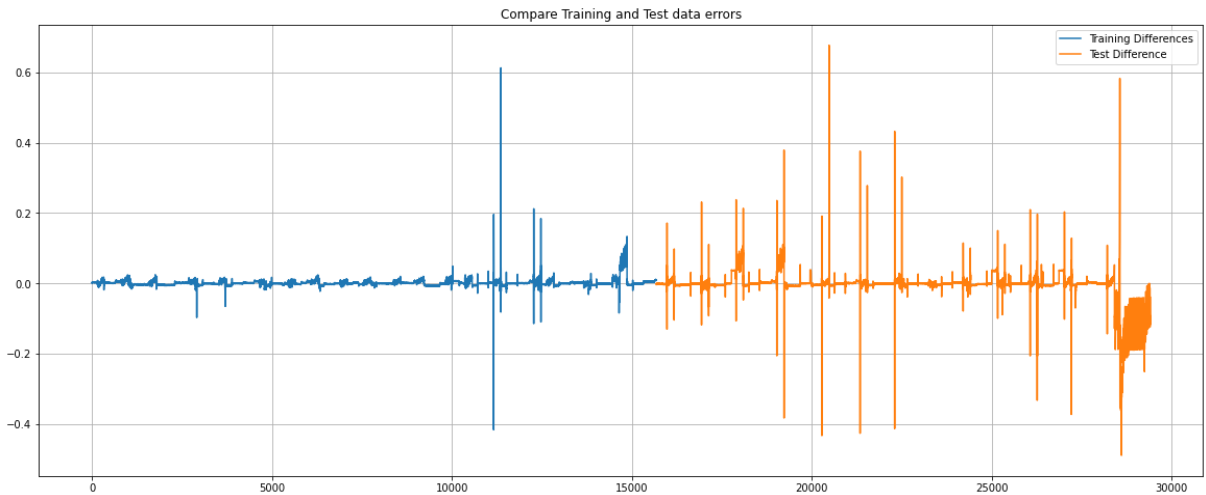

**Figure 8.** Prediction error of output vibration signal using Conv1D model and ADAM optimizer (5 Conv1D layers and over 2 million trainable weights over 500 epochs).

Using Tensorflow-Keras, we also produced results using an LSTM model which performs better than the Conv1D based model. Using the LSTM based model and ADAM optimizer, we can get close to the LM model that has a few dense layers, however you may need many more parameters and epochs. There is no certainty that the model would converge on the optimal solution using the TensorFlow ADAM optimizer due to the many hyper-parameter settings. Searching the hyper-parameter space is possible but it would be very costly. We can conclude that increasing the number of hidden units using the ADAM optimizer causes the model to approach the LM model result. A comparable fit is achieved at 512 hidden units or some 266k trainable parameters, see Figure 9. All model results confirm the presence of a number of robust anomalous features prior to the failure event.

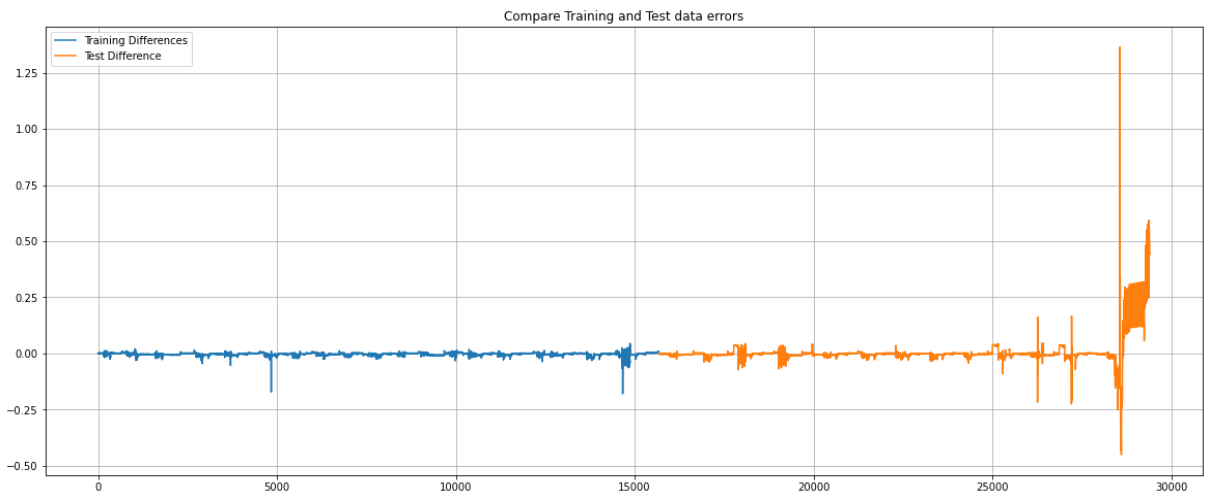

**Figure 9.** Prediction error of output vibration signal using 2-layer LSTM(128) sequential model and ADAM optimizer over 1000 epochs.

In terms of implementing LM on GPU (e.g. V100), we are able to get the LM algorithm to train with 1024 hidden units (over 1 million parameters) with GPU utilization rising to 11%. At 2048 hidden units (4.2 million parameters) GPU utilization rose to 35%. At 4096 hidden units (16 million parameters) GPU utilization rose to 70%. It topped out around 83% with 6000 hidden units (36 million parameters).

As we expected that the GPU utilization was low as the original problem size using just 10 hidden units was too small for a GPU. We had to increase the problem size, defined by the number of parameters to be estimated, by $10^5$ to see a heavy GPU usage.

We also saw only a small increase in the per epoch calculation time from 8 sec with 10 hidden neurons to just 14 sec with 2048 hidden neurons, then 28 sec with 4096 hidden neurons. Again, as we go to a bigger problem, we use the GPU more efficiently. This means that the compute time required to complete an epoch does not rise until we start to fill the GPU memory. As we can use the LM algorithm on multiple GPU's using the Horovod API to enable data parallelism, we expect that the use of LM optimizer can be expanded to relatively large-scale data problems.

In addition, we have been looking at PETSc TAO [6] as a possible highly scalable implementation of the LM algorithm, which would allow us to tackle very large problems. 'TAO' stands for Toolkit for Advanced Optimization; it focuses on algorithms for solving large-scale optimization problems on high-performance architectures. In TAO, there are many methods available for unconstrained, bound-constrained and generally constrained optimization, nonlinear least squares problems, variational inequalities and complementarity constraints. In [7], a LM regularizer is added to the BRGN routine, which converts BRGN into a Levenberg-Marquardt optimizer.

## 4. Summary

With a practical problem of anomaly detection from time series data with abrupt changes, we have investigated a number of combinations of NN architectures and optimizers. Using simple NN models, e.g. the dense layer only structure, we have found that the LM optimizer can perform very well while most other optimizers don't work well. Using more complicated NN structures and other optimizers such as the LSTM with ADAM optimizer, we need many more hidden neurons and epochs to match the performance of simple dense NN with LM optimizer. Many of the NN structures and optimizers don't work at all for the problem we have.

In applying LM optimizer to larger scale problems, we have also evaluated the use of LM optimizer in combination with GPU. We found that LM optimizer can still perform effectively provided the GPU has sufficient memory that matches the problem size. We have also determined that it is possible to utilize multiple GPU's which may be important for training on very large data sets. Currently, the implementation and application of the LM optimizer has some limitations in both the MATLAB and TensorFlow environments. MATLAB has not implemented the LM optimizer for deep learning models and GPU calculations, which limits the scale of the problem that can be addressed. The implementations of LM optimizer available on Github have not been vigorously tested, so performance can vary. For example, the Neupy package, where LM is built in, did not work for the problem presented above.

For promoting a wider use of the LM algorithm, we recommend that TensorFlow and Keras should fully implement the LM optimizer, and MATLAB should implement LM as a training optimization function for deep learning models.